\documentclass[conference]{IEEEtran}
\IEEEoverridecommandlockouts
\usepackage{cite}
\usepackage{amsmath,amssymb,amsfonts}
\usepackage{algorithmic}
\usepackage{graphicx}
\usepackage{textcomp}
\usepackage{xcolor}
\usepackage{algorithmic}
\usepackage{algorithm}
\usepackage{caption}
\usepackage{subcaption}
\usepackage{graphicx}
\usepackage{textcomp}
\usepackage{wrapfig}

\def\BibTeX{{\rm B\kern-.05em{\sc i\kern-.025em b}\kern-.08em
    T\kern-.1667em\lower.7ex\hbox{E}\kern-.125emX}}
\begin{document}

\title{Distributed multi-agent magnetic field norm SLAM with Gaussian processes\\

\thanks{
This publication is part of the project “Sensor Fusion For Indoor localization Using The Magnetic Field” with project number 18213 of the research program Veni which is (partly) financed by the Dutch Research Council (NWO). The experimental data in this publication was collected with the help of a customized app developed by Piet van Beek, Marnix Fetter, Bart de Jong, and Giel van der Weerd during their bachelor end project for Delft University of Technology.}
}


\author{\IEEEauthorblockN{Frida Viset$^{\star}$, Rudy Helmons$^{\dagger}$ and Manon Kok$^{\star}$}    \IEEEauthorblockA{$^{\star}$Delft Center for Systems and Control,$^{\dagger}$Maritime and Transport Technology, and Institute for Geoscience and Petroleum\\$^{\star}$ $^{\dagger}$Delft University of Technology, and $^{\dagger}$Norwegian University of Science and Technology \\{Email: \{F.M.Viset, R.L.J.Helmons, M.Kok-1\}@tudelft.nl}}}

\maketitle

\begin{abstract}

Accurately estimating the positions of multi-agent systems in indoor environments is challenging due to the lack of Global Navigation Satelite System (GNSS) signals. Noisy measurements of position and orientation can cause the integrated position estimate to drift without bound. Previous research has proposed using magnetic field simultaneous localization and mapping (SLAM) to compensate for position drift in a single agent. Here, we propose two novel algorithms that allow multiple agents to apply magnetic field SLAM using their own and other agents’ measurements.

Our first algorithm is a centralized approach that uses all measurements collected by all agents in a single extended Kalman filter. This algorithm simultaneously estimates the agents’ position and orientation and the magnetic field norm in a central unit that can communicate with all agents at all times. In cases where a central unit is not available, and there are communication drop-outs between agents, our second algorithm is a distributed approach that can be employed.

We tested both algorithms by estimating the position of magnetometers carried by three people in an optical motion capture lab with simulated odometry and simulated communication dropouts between agents. We show that both algorithms are able to compensate for drift in a case where single-agent SLAM is not. We also discuss the conditions for the estimate from our distributed algorithm to converge to the estimate from the centralized algorithm, both theoretically and experimentally.

Our experiments show that, for a communication drop-out rate of $80\%$, our proposed distributed algorithm, on average, provides a more accurate position estimate than single-agent SLAM. Finally, we demonstrate the drift-compensating abilities of our centralized algorithm on a real-life pedestrian localization problem with multiple agents moving inside a building.

\end{abstract}


\section{Introduction}

A wide range of research is being performed on multi-agent motion control and path planning algorithms~\cite{oh_survey_2015}. For most motion control algorithms, it is crucial for each agent to know its own position~\cite{egerstedt_formation_2001, sandeep_decentralized_2006}. Collaborative pedestrian navigation can be useful for example for rescue missions or law enforcement applications~\cite{fankhauser_collaborative_2016}. Indoors, Global Navigation Satellite System (GNSS) signal availability is limited and prone to errors~\cite{puricer_technical_2007}. Current indoor navigation systems therefore often rely on integrating measurements of the change in position and orientation. For autonomous navigation in GNSS-denied environments where there are no previously deployed beacons or other structure supporting navigation, measurements of the change in position and orientation are often available from for example inertial sensors, wheel encoders or visual-inertial odometry~\cite{mohamed_survey_2019}. Integrating measurements of change in position and orientation (odometry) gives accumulated position estimation errors (drift) that can increase without an upper bound~\cite{woodman_introduction_2007}. 

To compensate for odometry drift, multi-agent simultaneous localization and mapping (SLAM) algorithms for navigation in GNSS-denied environments based on visual information have been widely studied~\cite{zou_collaborative_2019}. Visual SLAM can in some applications be infeasible or prone to error due to privacy concerns, varying light conditions, or lack of distinguishable features or landmarks~\cite{tourani_visual_2022}. 

\begin{figure}
    \centering
\includegraphics[trim={7cm 5cm 6cm 4cm},clip,width=0.45\textwidth]{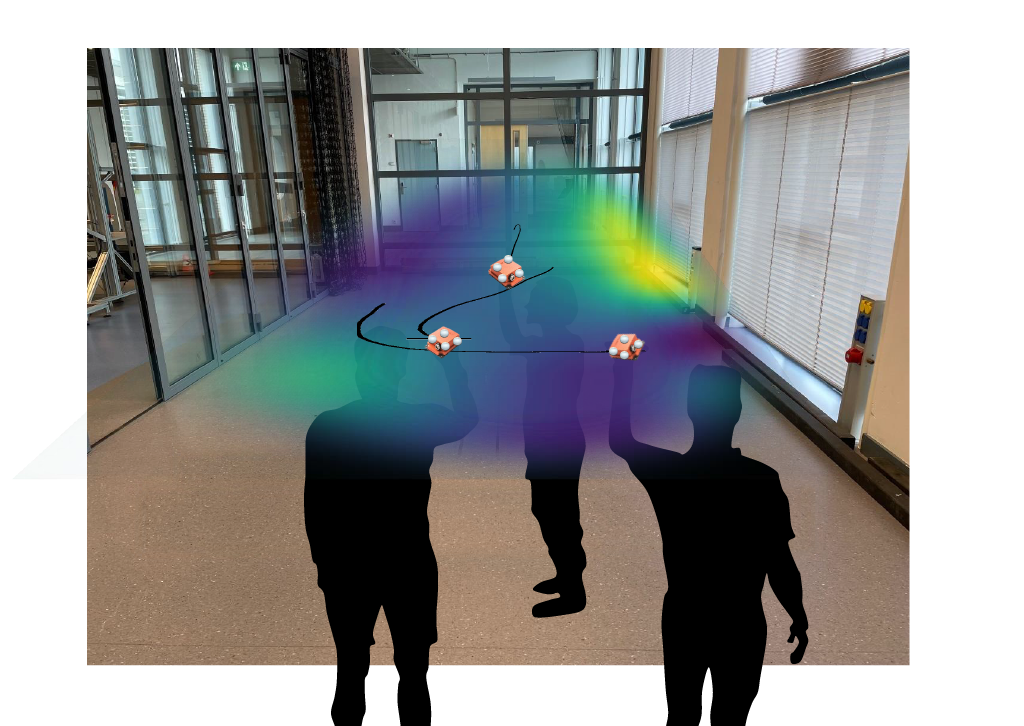}
    \caption{Estimated magnetic field map and trajectories of multiple agents based on measurements from magnetometers carried by three people. The position of the magnetometer was recorded in an optical motion capture lab. The color of the map reflects the magnitude of the magnetic field norm, while the opacity of the overlaid map is inversely proportional to the marginal variance of the estimate.}
    \label{fig:exp_setup}
\end{figure}

For several single-agent navigation tasks, magnetic field SLAM has been proposed and demonstrated to compensate for drift in the position estimate ~\cite{osman_indoor_2022, vallivaara_magnetic_2011, vallivaara_simultaneous_2010, coulin_online_2022, kok_scalable_2018, robertson_simultaneous_2009, robertson_simultaneous_2013,jung_indoor_2015}. The magnetic field indoors is affected by structural metallic elements~\cite{storms_magnetic_2010-1}. In Figure~\ref{fig:exp_setup}, an example of the magnetic field norm variations that can be found indoors is displayed. The indoor magnetic field typically has significant spatial variations and stays constant over time~\cite{ouyang_survey_2022, frassl_magnetic_2013}. To simultaneously create and use a map of the magnetic field, most approaches use a nonlinear stochastic interpolation scheme to learn the magnetic field online based on measurements. A stochastic interpolation scheme that also gives an uncertainty measure on the predictions in every location of the map is Gaussian process regression. Several of the previous works into magnetic field SLAM use reduced-rank Gaussian process regression approximated with Hilbert space basis functions so the computational complexity does not scale with the number of measurements~\cite{osman_indoor_2022,vallivaara_magnetic_2011, kok_scalable_2018, viset_extended_2022}. 

The contribution of this paper is twofold. The first contribution is an algorithm that uses all information measured by multiple agents to perform magnetic field norm SLAM online with an extended Kalman filter (EKF). This EKF is obtained by augmenting the state-space of the EKF for magnetic field SLAM in~\cite{viset_extended_2022} to contain the poses of multiple agents as opposed to just a single agent. We denote this as the centralized algorithm, as it is an algorithm that can be executed in a centralized station that receives all measurements made by all agents. Multi-agent systems do not always have access to a centralized control unit. Our second contribution is therefore a distributed version of the algorithm, where each agent uses information shared in communication between the agents to collaboratively approximate the output of the centralized algorithm. To implement the centralized EKF as a decentralized EKF, we use an approach closely related to the decentralized Kalman filter described in~\cite{olfati-saber_distributed_2005}. To the best of the author's knowledge, this is the first proposed algorithm for distributed multi-agent magnetic field SLAM with Gaussian process regression. 

\section{Connections to previous work}

Previous work has applied average consensus to achieve distributed reduced-rank Gaussian process regression using measurements from multiple agents~\cite{pillonetto_distributed_2019,jang_multi-robot_2020}. Recursive stochastic least squares correspond to applying repeated Kalman filter measurement updates~\cite{sarkka_bayesian_2013}. Magnetic field SLAM with an extended Kalman filter uses both a dynamic update and a measurement update at each timestep to jointly estimate the magnetic field map and the pose of a single agent~\cite{viset_extended_2022}. Previous work has also demonstrated that Kalman filters with both measurement updates and dynamic updates can be implemented for multiple agents distributively with embedded consensus filters~\cite{olfati-saber_distributed_2005}. The distributed implementation in~\cite{olfati-saber_distributed_2005} is implemented by solving two consensus problems at each time step, one in the dynamic update and one in the measurement update. We also implement the distributed EKF by solving these two consensus problems at each time step. For both our distributed EKF and for the distributed Kalman filter in~\cite{olfati-saber_distributed_2005}, even if each average consensus problem has not converged, the intermittent result is an approximation of the centralized solution~\cite{xiao_scheme_2005}. 

Unlike previous work into extended Kalman filtering for magnetic field SLAM, we execute the measurement update on the information form. This allows for the measurement update to be implemented distributively by executing the average consensus algorithm at each timestep. Performing the measurement update for magnetic field SLAM on information form is closely related to the execution of the measurement updates on information form for magnetic field mapping proposed by~\cite{jang_multi-robot_2020}. The main difference between our work and the estimation algorithm presented in~\cite{jang_multi-robot_2020} is that we jointly and distributively estimate the pose of the agents and the map, while~\cite{jang_multi-robot_2020} only estimates the map. The main difference between our work and~\cite{viset_extended_2022} and~\cite{viset_magnetic_2021} is that we perform magnetic field SLAM for several agents instead of just one and that we propose a distributed algorithm for doing so. An additional difference between our work and the work presented in~\cite{viset_extended_2022} is that we for simplicity consider only the magnetic field norm instead of the three-component magnetic field.


\section{Model}

We assume that each individual agent has access to noisy odometry measurements, according to a model we describe in Section~\ref{sec:dynamic_model}. We also assume that each agent carries a magnetometer capable of measuring the magnetic field norm. In Section~\ref{sec:measurement_model} we give the measurement model for the magnetometer and the model of the magnetic field norm that we use to apply Gaussian process regression to learn the magnetic field map.

\subsection{Dynamic model}\label{sec:dynamic_model}

We estimate the position of a set of $m$ agents indexed as $i={1,\hdots,m}$. The position and orientation of each agent at each timestep $t$ are denoted by the vector $p_{i,t}$ and the unit quaternion $q_{i,t}$ respectively. The quaternion is defined as the orientation from the world frame to the body frame. The body frame has its origin in the IMU's center of mass, and its axes are aligned with the accelerometer sensor axes. The world frame is defined as the stationary inertial frame that shares its origin with the body frame at time $t=0$, where the gravity field is aligned with the negative z-axis, and the initial yaw-angle between the body and world-frame at $t=0$ is zero. The position is given in the world frame.

We assume that each agent has access to noisy measurements $\Delta p_{i,t}$ of the change in their position and $\Delta q_{i,t}$ of the change in their orientation from sensors mounted in the body frame.  The noisy measurements are defined such that
\begin{subequations}
\begin{align}
    p_{i,t+1}=&\:p_{i,t}+R(q_{i,t})(\Delta p_{i,t} + e_{i,\text{p},t}), \label{eq:dynamic_model1}\\
    q_{i,t+1}=&\:q_{i,t}\odot\exp_{\text{q}}(\Delta q_{i,t})\odot\exp_{\text{q}}(e_{i,\text{q},t}),\label{eq:dynamic_model2}\\
     [e_{i,\text{p},t}^\top, e_{i,\text{q},t}^\top ]^\top\sim&\:\:\mathcal{N}(0,\Sigma), \label{eq:dynamic_model3}
\end{align}
\end{subequations}
where $e_{i,\text{p},t}$ is a measurement noise of the change in position, $e_{i,\text{q},t}$ is a measurement noise of the change in orientation, and where $\Sigma$ is a known noise covariance, $\odot$ is the quaternion product, and $\exp_{\text{q}}$ is the operator that maps an axis-angle orientation deviation to a quaternion, defined as in the odometry model in~\cite{kok_scalable_2018}, and where ${R}(\cdot)$ is an operator transforming a unit quaternion to a rotation, defined as in the odometry model in~\cite{viset_magnetic_2021}. Note that we assume the odometry covariance is the same for all agents. 

\subsection{Measurement model}\label{sec:measurement_model}

We assume that each agent $i$ has access to a continuous stream of measurements from the magnetic field norm in their current position $p_{i,t}$, according to
\begin{equation}
    y_{i,t}=f(p_{i,t})+e_{i,t},\qquad e_{i,t}\sim\mathcal{N}(0,\sigma_\text{y}^2),\label{eq:measurement_model}
\end{equation}
where $y_{i,t}$ is the measurement from agent $i$ at time $t$, $f:\mathbb{R}^3\rightarrow\mathbb{R}$ is a function that maps the position to the magnetic field norm, and $e_{i,t}$ is the measurement noise with a covariance $\sigma_\text{y}^2$. We model the function $f$ as a stationary Gaussian process according to
\begin{equation}\label{eq:GP_prior}
    f~\sim\mathcal{GP}(0,\kappa_{\text{SE}}(\cdot,\cdot)),
\end{equation}
with a squared exponential kernel
\begin{equation}
    \kappa_{\text{SE}}(x,x')=\sigma_{\text{SE}}^2 \exp \left (-\frac{\|x-x'\|_2}{2l_{\text{SE}}^2}\right ),
\end{equation}
where $\sigma_{\text{SE}}^2$ and $l_{\text{SE}}$ are hyperparameters denoting the variance and lengthscale of the magnetic field norm nonlinearities, respectively~\cite{viset_magnetic_2021}. We use the same basis functions as~\cite{viset_extended_2022} and~\cite{viset_magnetic_2021} to approximate the Gaussian process regression. The basis functions are defined in Appendix~\ref{app:basis_functions}. 

\subsection{Communication graph}

We assume that the agents have the possibility to send and receive $N_c$ messages to all other agents two times at each timestep $t$, once for the dynamic update and once for the measurement update. 

We model the communication graph at time $t$ and communication step $t_c$ as an undirected graph $\mathcal{G}(t,t_c)=(\mathcal{E}(t,t_c),\mathcal{V})$ where $\mathcal{E}(t,t_c)\subset \{\{i,j\}|i,j\in \mathcal{V}\}$ denote the set of active communication edges at time $t$ between the set $\mathcal{V}=\{1,\cdots,m\}$ of all agents. We assume the probability for two agents to be able to communicate at any timestep $t$ at communication step $t_c$ to be $1-\alpha$, where $\alpha$ is the probability of communication failure. We will refer to $\alpha$ as the communication failure rate or the dropout rate in the remainder of this paper. We denote the communication step at each timestep by the index $t_c$, where $t_c=1,\hdots, 2N_c$.

\section{Centralized EKF for multi-agent magnetic field SLAM}\label{sec:problem_formulation}

Following the approach of~\cite{viset_extended_2022}, we parameterize our system in terms of an error state $\xi_t$ linearised about the prior beliefs of the position of agents $i=1,\hdots,m$ denoted $\tilde{p}_{i,t|t-1}$, the prior beliefs of the orientation of agent $i=1,\hdots,m$ denoted $\tilde{q}_{i,t|t-1}$ and the prior belief of the map denoted $\tilde{w}_{i,t|t-1}$. The error state $\xi_t$ is defined as
\begin{align}\label{eq:def_error_state_filtered}
    {\xi}_{t}&=[\delta_{1,t}^\top\quad\eta_{1,t}^\top \quad \cdots\quad \delta_{m,t}^\top\quad \eta_{m,t}^\top\quad \nu_{t}^\top]^\top,
\end{align}
where $\delta_{i,t}=p_{i,t}-\tilde{p}_{i,t|t-1}$ denotes the position estimation error, $\nu_{t,i}=w-\tilde{w}_{i,t|t-1}$ denotes the magnetic field state estimation error, and $\eta_{i,t}$ denotes the orientation estimation error parameterized as an axis-angle deviation according to
\begin{align}
    q_{i,t}=&\exp_{\text{q}}(\eta_{i,t})\odot\tilde{q}_{i,t|t-1}.
\end{align}
For simplicity, we assume that the initial position and orientation of all agents are known. The initial error state is then distributed as $\xi_{0}\sim\mathcal{N}(0,P_{0|0})$, where $P_{0|0}$ is given by
\begin{equation}
    P_{0|0}=\begin{bmatrix}
0 & \cdots & 0 & 0\\ 
\vdots & \ddots & \vdots & \vdots \\ 
0 & \cdots & 0 & 0 \\ 
0 & \cdots & 0 & \Lambda, \\ 
\end{bmatrix}.\label{eq:P_init}
\end{equation}
with $\Lambda$ defined in~\eqref{eq:red_rank_prior}.

\subsection{Dynamic update}
The posterior linearisation point is propagated to a prior linearisation point by applying the dynamic model in~\eqref{eq:dynamic_model1}-\eqref{eq:dynamic_model3} through the update
\begin{subequations}
\begin{align}
\tilde{q}_{i,t+1|t}=&\tilde{q}_{i,t|t}\odot\exp_{\text{q}}(\Delta q_{i,t}),  &i=1,\hdots,m\label{eq:dynamic_relin1}\\ \tilde{p}_{i,t+1|t}=&\tilde{p}_{i,t|t}+R(\tilde{q}_{i,t|t})\Delta p_{i,t}, &i=1,\hdots,m\label{eq:dynamic_relin2}\\
    \tilde{w}_{t+1|t}=&\tilde{w}_{t|t}\label{eq:dynamic_relin3}.
\end{align}
\end{subequations}
The centralized dynamic update is defined as
\begin{equation}\label{eq:dynamic_covariance_update}
    P_{t+1|t}=F_tP_{t|t}F_{t}^\top+Q,
\end{equation}
where $Q$ is given by
\begin{equation}
    Q=\begin{bmatrix}
\Sigma & \hdots  & 0 & 0\\
\vdots  & \ddots & \vdots & \vdots\\
0 & \hdots   & \Sigma & 0\\
0 & \hdots & 0 & 0
\end{bmatrix}
\end{equation}
and $F_t$ is defined as
\begin{equation}
F_t=\begin{bmatrix}
F_{1,t} & \hdots & 0 & 0\\ 
\vdots & \ddots & \vdots & \vdots\\ 
0 & \hdots & F_{m,t} & 0\\ 
0 & \hdots & 0 & I
\end{bmatrix},
\end{equation}
where the matrix $F_{j,t}$ is given by
\begin{equation}
    F_{j,t}=\begin{bmatrix}
I  & R(\tilde{q}_{j,t|t})[\Delta p_{j,t}\times] \\ 0 & I 
\end{bmatrix},
\end{equation}
and where $[\Delta p_{i,t}\times]$ is defined as the skew-symmetric matrix such that $[\Delta p_{i,t}\times] u=\Delta p_{i,t}\times u$ gives the cross-product between $\Delta p_{i,t}$ and a vector $u\in\mathbb{R}^3$.

\subsection{Measurement update}

The measurement update is performed by linearising the measurement model in~\eqref{eq:measurement_model} about the prior linearisation point with respect to the error state $\xi_t$. We let the information vector $\iota_{t|t-1}$ and information matrix $\mathcal{I}_{t|t-1}$ denote the information form of the state estimate $\hat{\xi}_{t|t-1}$ and the corresponding covariance $P_{t|t-1}^{-1}$, according to
\begin{align}
    \iota_{t|t-1}=&P_{t|t-1}^{-1}\hat{\xi}_{t|t-1}=0,\\
    \mathcal{I}_{t|t-1}=&P_{t|t-1}^{-1}.\label{eq:standard_to_information}
\end{align}
The Kalman filter measurement update can then be expressed as an update of the information matrix and information vector as
\begin{subequations}
\begin{align}
 \mathcal{I}_{t|t}=&\mathcal{I}_{t|t-1}+\sum_{i=1}^{m}\frac{1}{\sigma_\text{y}^2}H_{i,t}H_{i,t}^\top,\label{eq:measurement_update1}\\
\iota_{t|t}=&\iota_{t|t-1}+\sum_{i=1}^{m}\frac{1}{\sigma_\text{y}^2}H_{i,t}(y_{i,t}-\Phi(\tilde{p}_{i,t-1})^\top\tilde{w}_{t|t-1}),\label{eq:measurement_update2}
\end{align}
\end{subequations}
with 
\begin{equation}
\begin{split}
    H_{i,t}=&[0_{1\times 6(i-1)},(\nabla\Phi(\tilde{p}_{i,t|t-1})\tilde{w}_{t|t-1})^\top, \\
    &0_{1\times (3+6(m-i))},(\Phi(\tilde{p}_{i,t|t-1}))^\top]^\top.
\end{split}
\end{equation}
The posterior error state estimate and covariance are given by
\begin{align}\hat{\xi}_{t|t}=\mathcal{I}_{t|t}^{-1}\iota_{t|t}, \qquad
P_{t|t}=\mathcal{I}_{t|t}^{-1}.\label{eq:information_to_standard}
\end{align}
The posterior linearisation point can then be calculated by propagating the estimated error state to the prior linearisation point according to
\begin{subequations}
\begin{align}
    \tilde{p}_{i,t|t}=&\tilde{p}_{i,t|t-1}+\hat{\delta}_{i,t|t},  &i=1,\hdots,m, \label{eq:relinearise1}\\
    \tilde{q}_{i,t|t}=&\exp_{\text{q}}(\hat{\eta}_{i,t|t})\odot\tilde{q}_{i,t|t-1},  &i=1,\hdots,m, \label{eq:relinearise2}\\
    \tilde{w}_{t|t}=&\tilde{w}_{t|t-1}+\hat{\nu}_{t|t}.\label{eq:relinearise3}
\end{align}
\end{subequations}
Recursively applying the dynamic update and measurement update results in the centralized EKF for multi-agent magnetic field SLAM, as described in Algorithm~1.

\begin{algorithm}[!t]
 \caption{Centralized EKF for multi-agent magnetic field SLAM}\label{alg:central}
 \begin{algorithmic}[1]
 \renewcommand{\algorithmicrequire}{{Input:}}
 \renewcommand{\algorithmicensure}{{Output:}}
 \REQUIRE $\left \{\{\Delta p_{i,t}, \Delta q_{i,t}, {y}_{i,t} \}_{t=1}^N \right \}_{i=1}^{m}$
 \\ \vspace{1mm}
 \ENSURE  $\left \{\left \{{\tilde{p}} _{i,t|t} ,\:\tilde{q} _{i,t|t},\:\tilde{w}_{t|t} \right\}_{t=1}^N \right \}_{i=1}^{m}$
 \\ \vspace{1mm} \textit{Initialization}: $\tilde{p}_{i,0|0}={0}_{3\times1}$, $\tilde{{q}}_{i,0|0}={q}_{0,i}$, $\tilde{w}_{0|0}=0_{M\times 1}$,~\eqref{eq:P_init}
  \FOR {$t = 1$ to $N$}
  \STATE Dynamic update according to~\eqref{eq:dynamic_relin1}, \eqref{eq:dynamic_relin2}, \eqref{eq:dynamic_relin3} and~\eqref{eq:dynamic_covariance_update}.
  \STATE Measurement update according to~\eqref{eq:standard_to_information}, \eqref{eq:measurement_update1}, \eqref{eq:measurement_update2} and~\eqref{eq:information_to_standard}. Relinearization according to~\eqref{eq:relinearise1}, \eqref{eq:relinearise2} and~\eqref{eq:relinearise3}.
  \ENDFOR
 \end{algorithmic}
 \end{algorithm}

\section{Distributed multi-agent EKF for magnetic field SLAM}

We denote agent $i$'s approximation of a centralized term by including a superscript $(i)$ on the approximated term. The initial posterior linearisation points are known and given as $\tilde{p}_{i,0|0}^{{(i)}}=p_{i,0}$,  $\tilde{q}_{i,0|0}^{{(i)}}=q_{i,0}$ and $\tilde{w}_{0|0}^{{(i)}}=0$. As in the centralized filter, we assume that the initial error centralized error state $\hat{\xi}_{t|t}^{(i)}=0$ and the initial centralized covariance $P_{0|0}^{(i)}=P_{0|0}$ are both known.

\subsection{Dynamic update}
In the case where each agent only has access to their own measurements, the posterior linearisation point of each agent can be propagated to a prior linearisation point through the dynamic model in the same way as for the centralized EKF, using~\eqref{eq:dynamic_covariance_update}. The matrix $F_t$ cannot be computed directly by any agent as each term $F_{j,t}$ contains the odometry measurement $\Delta p_{j,t}$ which is only available to agent $j$. The matrix $F_t$ can however be approximated by the network as a whole through average consensus, if each agent initializes their belief about the matrix $F_t^{(i)}$ according to
\begin{equation}\label{eq:dyn_up_avg_prep1}
    F_t^{(i)}=\begin{bmatrix}
F_{1,t}^{(i)} & \hdots & 0 & 0\\ 
\vdots & \ddots & \vdots & \vdots\\ 
0 & \hdots & F_{m,t}^{(i)} & 0\\ 
0 & \hdots & 0 & I
\end{bmatrix},
\end{equation}
where the term $F_{j,t}^{(i)}$ is defined according to
\begin{equation}\label{eq:dyn_up_avg_prep2}
    F_{j,t}^{(i)}=\left\{\begin{matrix}
mF_{j,t}-(m-1)I, & j=i\\
I, & j\neq i 
\end{matrix}\right.
\end{equation}
The average of all the terms $\{F_{t}^{(i)}\}_{i=1}^{m}$ is $F_{t}$, so applying average consensus according to
\begin{equation}\label{eq:dyn_avg_exec}
F_t^{(i)}\leftarrow\sum_{j=1}^{m}W_{i,j}(t,t_c)F_t^{(j)},
\end{equation}
where the weights $W_{i,j}(t,t_c)$ are defined as in~\cite{xiao_scheme_2005} as
\begin{equation}
W_{i,j}(t,t_c)=\left\{\begin{matrix}
\frac{1}{m}, & {i,j}\in\mathcal{E}(t,t_c)\\ 
1-\frac{d_{i}(t,t_c)}{m}, & i=j\\ 
0, & \text{otherwise}
\end{matrix}\right.,
\end{equation}
and where $d_{i}(t,t_c)$ are the number of edges to node $i$ in the communication graph $\mathcal{E}(t,t_c)$ at timestep $t$, for $t_c=1,\hdots,N_c$ causes $F_{t}^{(i)}$ to converge to $F_t$ as $N_c\rightarrow\infty$~\cite{xiao_scheme_2005}. As we only apply a finite amount of average consensus steps $N_c$, we use $F_t^{(i)}$ at time $N_c$ as an approximation in the dynamic update of the covariance.

\subsection{Measurement update}
The measurement update can be carried out in a distributed manner by first letting each agent update its belief about the information vector according to
\begin{subequations}
\begin{align}
        \mathcal{I}_{t|t}^{(i)}=&\mathcal{I}_{t|t-1}^{(i)}+m\frac{1}{\sigma_\text{y}^2}H_{i,t}H_{i,t}^\top,\label{eq:dist_meas_update1}\\
\iota_{t|t}^{(i)}=&\iota_{t|t-1}^{(i)}+m\frac{1}{\sigma_\text{y}^2}H_{i,t}(y_{i,t}-\Phi(\tilde{p}_{i,t-1}^{(i)})^\top\tilde{w}_{t|t-1}^{(i)}),\label{eq:dist_meas_update2}
\end{align}
\end{subequations}
and then carry out average consensus across the network on the resulting information matrix and information vector, according to
\begin{subequations}
\begin{align}
    \iota_{t|t}^{(i)}\leftarrow\sum_{j=1}^m W_{i,j}(t,t_c)\iota_{t|t}^{(j)}\label{eq:dist_avg_exec1}\\
    \mathcal{I}_{t|t}^{(i)}\leftarrow\sum_{j=1}^m W_{i,j}(t,t_c)\mathcal{I}_{t|t}^{(j)}\label{eq:dist_avg_exec2}
\end{align}
\end{subequations}
The result will then converge to the information matrix and information vector obtained by~\eqref{eq:measurement_update1}-~\eqref{eq:measurement_update2} as the number of communication steps goes to infinity. We use the output from the average consensus procedure as an approximation to the centralized information matrix in each agent. Each agent can therefore update their own linearization point locally by using the same update as the centralized EKF in~\eqref{eq:relinearise1}-\eqref{eq:relinearise3}. When there is no communication failure, the approximation will be exactly equal to the centralized solution even with $N_c=1$. Recursively applying the dynamic update and the measurement update gives the Distributed EKF for multi-agent magnetic field SLAM described in Algorithm~2. If all $m$ agents are running Algorithm~2, the multi-agent system will collaboratively approximate the centralized estimate of Algorithm~1.

\begin{algorithm}[!t]
 \caption{Distributed EKF for multi-agent magnetic field SLAM for agent $i$}\label{alg:distributed}
 \begin{algorithmic}[1]
 \renewcommand{\algorithmicrequire}{{Input:}}
 \renewcommand{\algorithmicensure}{{Output:}}
 \REQUIRE $\{\{\Delta p_{i,t}, \Delta q_{i,t}, {y}_{i,t} \}_{t=1}^N\}_{i=1}^m $
 \\ \vspace{1mm} 
 \ENSURE  $\left\{ \left \{{\tilde{p}} _{i,t|t}^{(i)} ,\:\tilde{q} _{i,t|t}^{(i)},\:\tilde{w}_{t|t}^{(i)} \right\}_{t=1}^N \right\}_{i=1}^m$
 \\ \vspace{1mm}  \textit{Initialization}: $\tilde{p}_{i,0|0}^{(i)}={0}_{3\times1}$, $\tilde{{q}}_{i,0|0}^{(i)}={q}_0$, $\tilde{w}_{0|0}^{(i)}=0_{M\times 1}$,  ~\eqref{eq:P_init}
  \FOR {$t = 1$ to $N$}
  \STATE \textbf{Dynamic update:} Perform average consensus according to~\eqref{eq:dyn_up_avg_prep1},~\eqref{eq:dyn_up_avg_prep2} and~\eqref{eq:dyn_avg_exec}. Then, propagate own belief of own state according to~\eqref{eq:dynamic_relin1}, \eqref{eq:dynamic_relin2}, \eqref{eq:dynamic_relin3} and~\eqref{eq:dynamic_covariance_update}, using output terms from average consensus.
  \STATE \textbf{Measurement update:} according to~\eqref{eq:dist_meas_update1}, \eqref{eq:dist_meas_update2}. Average consensus according to~\eqref{eq:dist_avg_exec1} and~\eqref{eq:dist_avg_exec2}. Relinearization according to~\eqref{eq:relinearise1}, \eqref{eq:relinearise2} and~\eqref{eq:relinearise3}.
  \ENDFOR
 \end{algorithmic}
\end{algorithm}

\section{Results}

\subsection{Comparison of Algorithm~1 to Single-Agent SLAM}

\begin{figure*}
    \hspace{15pt}
     \begin{subfigure}[]{0.18\textwidth}
         \includegraphics[width=\textwidth]{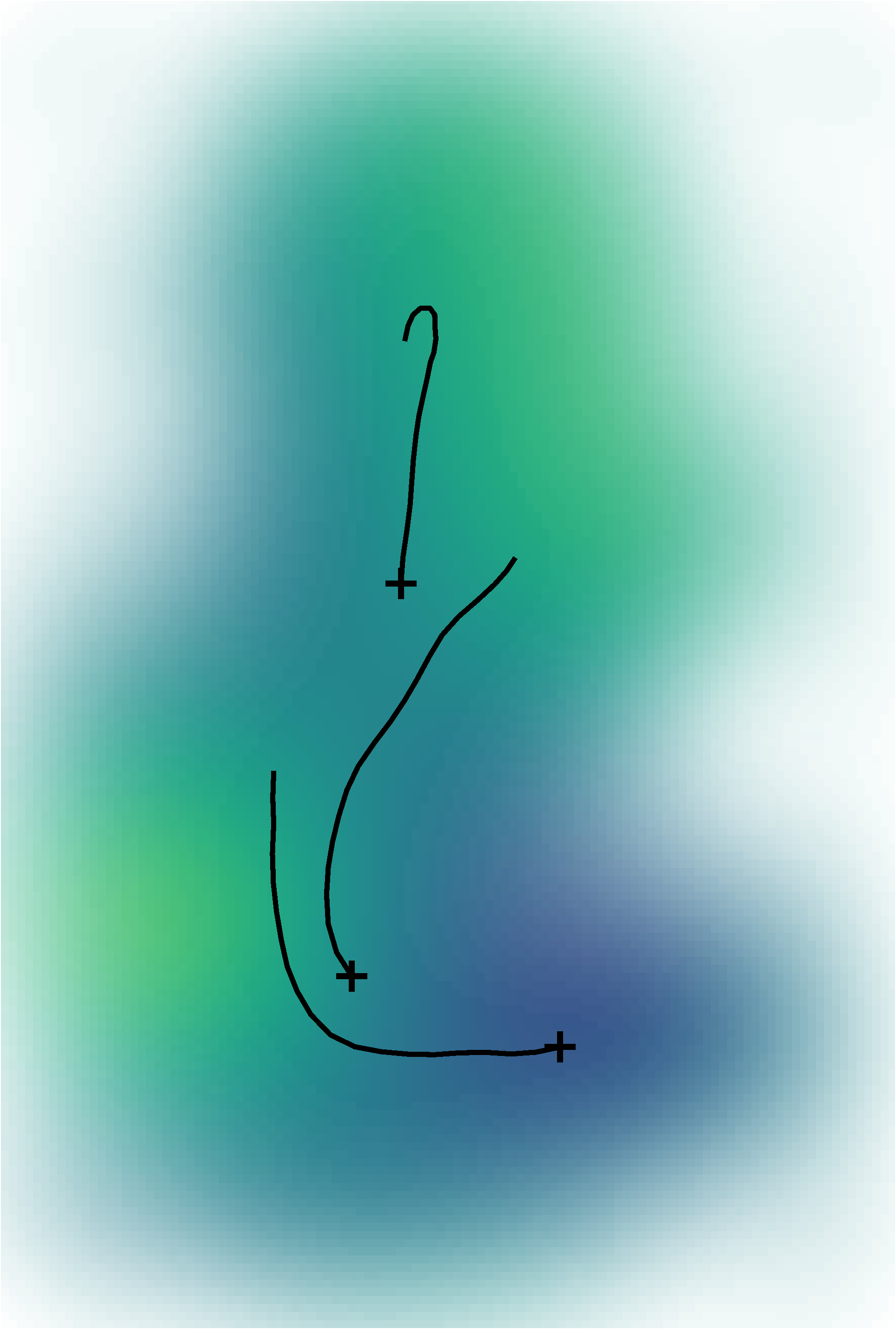} 
         \label{fig:time2seconds}
     \caption{$t$ = 2 seconds}
     \end{subfigure}
         \hspace{15pt}
     \begin{subfigure}[]{0.18\textwidth}
         \includegraphics[width=\textwidth]{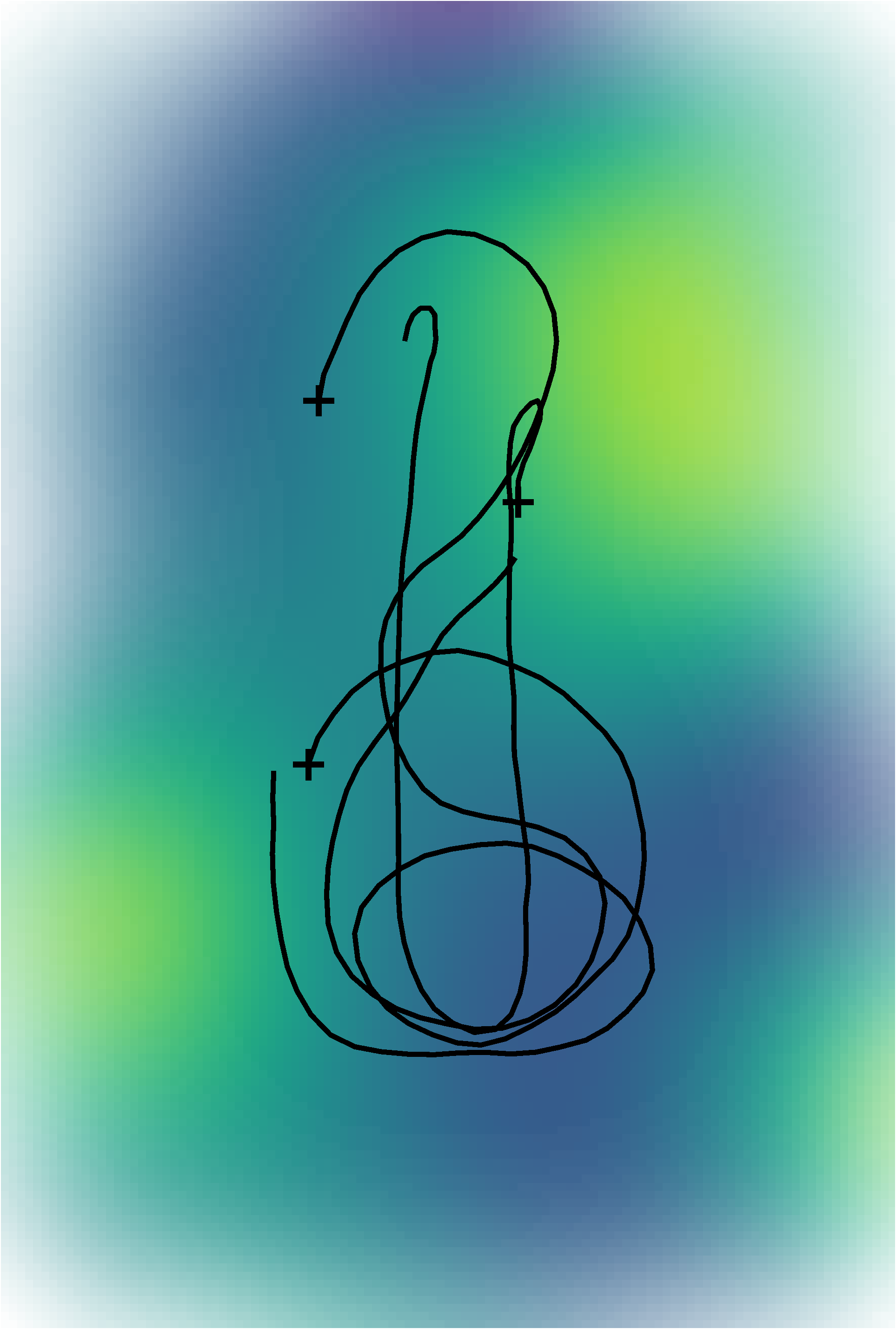}
         \label{fig:time7seconds}
     \caption{$t$ = 7 seconds}
     \end{subfigure}
         \hspace{15pt}
    \begin{subfigure}[]{0.18\textwidth}
         \includegraphics[width=\textwidth]{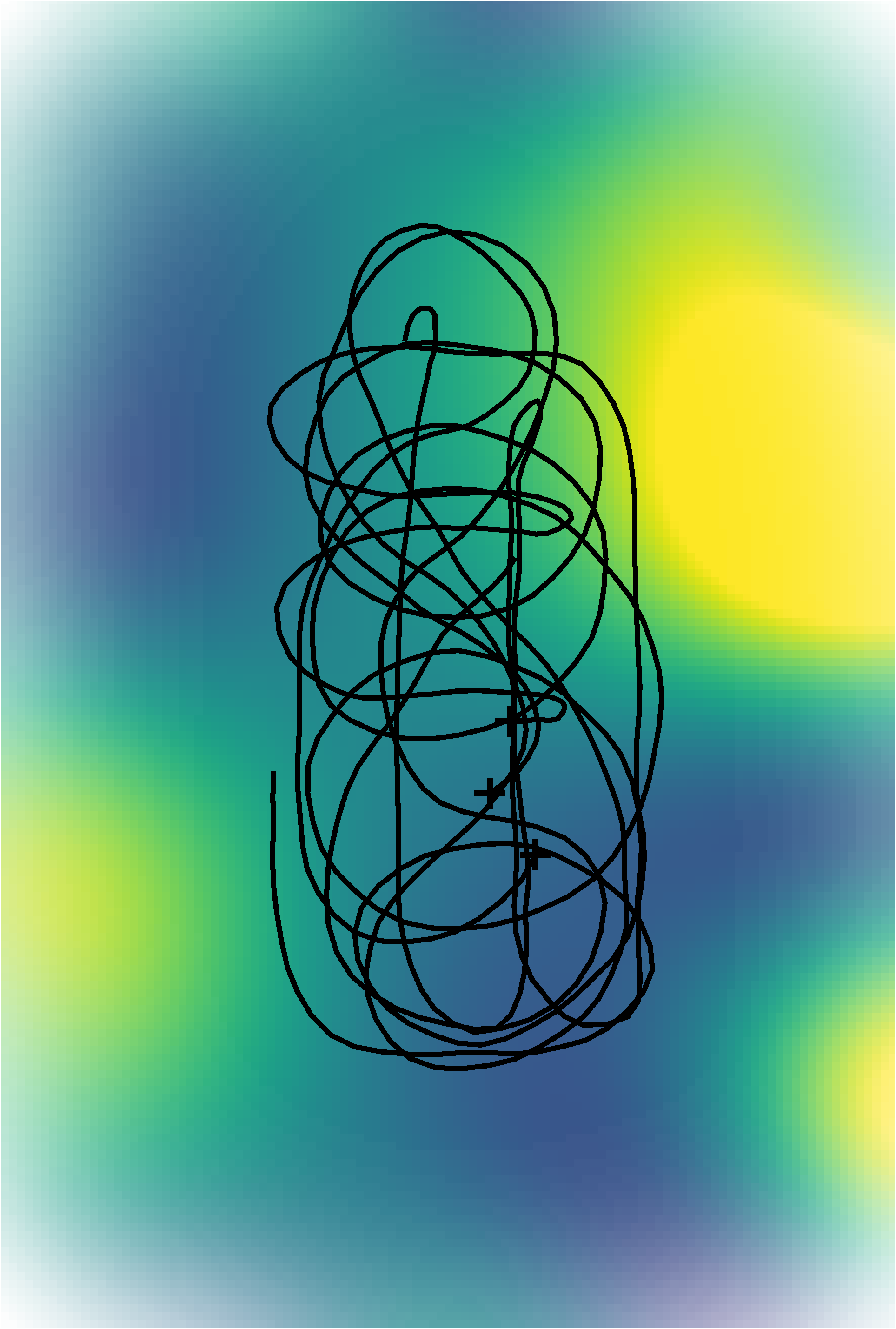}
         \label{fig:time20seconds}
     \caption{$t$ = 20 seconds}
     \end{subfigure}
         \hspace{15pt}
     \begin{subfigure}[]{0.18\textwidth}
         \includegraphics[width=\textwidth]{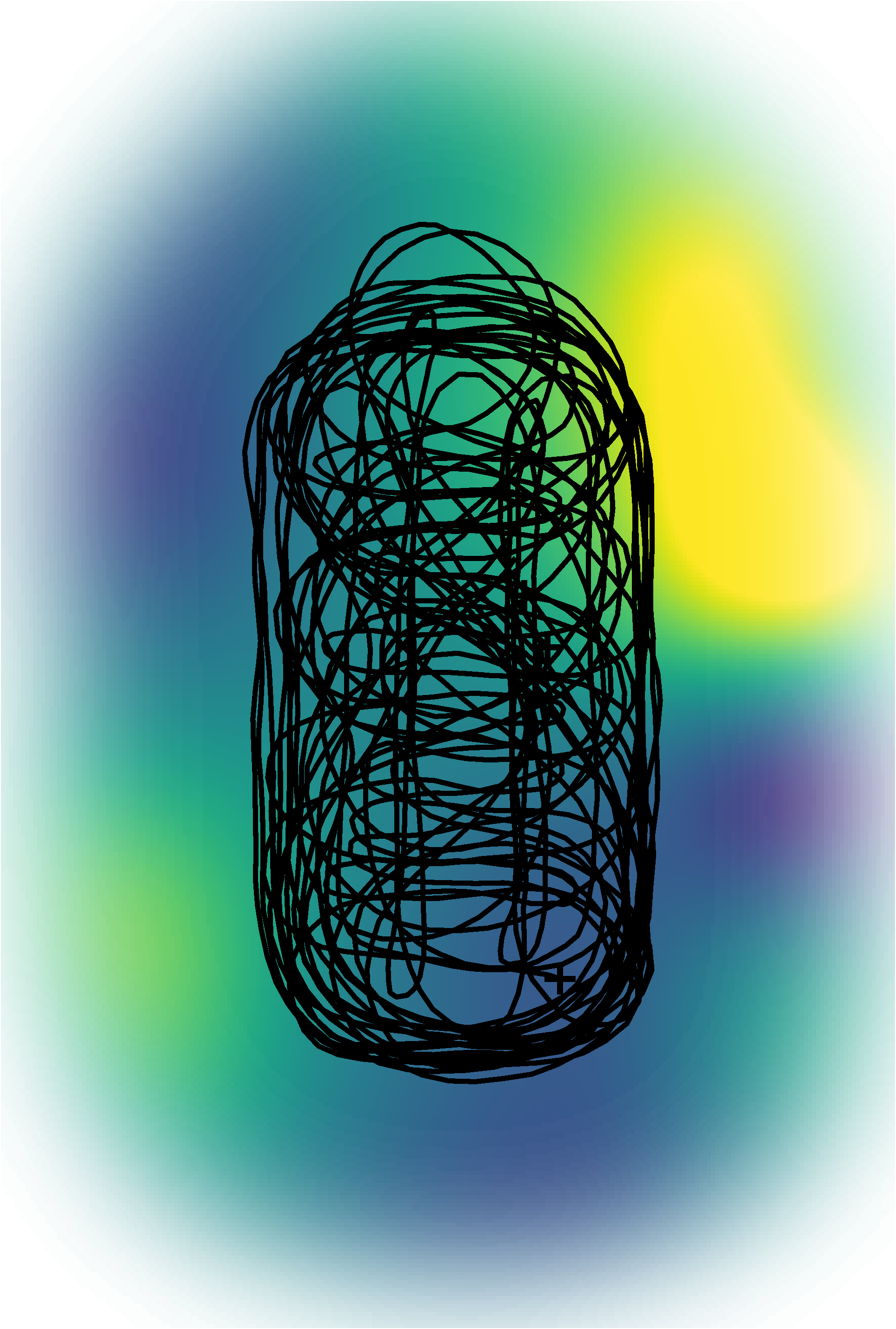}
         \label{fig:time80secons}
     \caption{$t$ = 80 seconds}
     \end{subfigure}
     \hspace{15pt}
     \begin{subfigure}[]{0.06\textwidth}
         \centering
         \includegraphics[width=\textwidth]{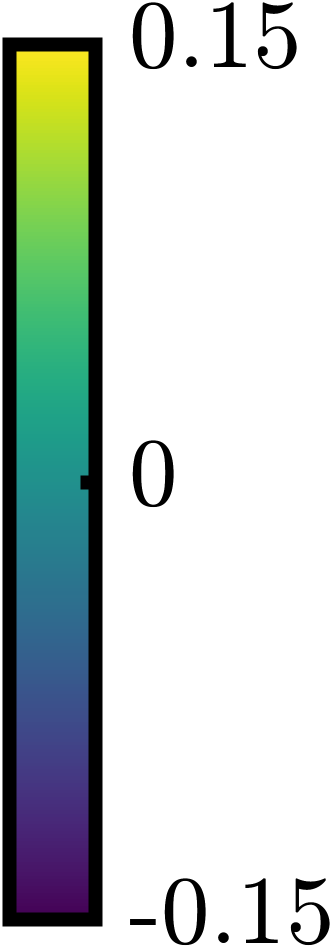} 
         \label{fig:agentslegends}
     \end{subfigure}
        \caption{Learned magnetic field map using Algorithm 1. The intensity of the learned magnetic field norm is indicated by the color, while the marginal variance of the magnetic field map is inversely proportional to the opacity. The estimated trajectories of the agents are indicated with black lines, and the current positions at each time are indicated with black crosses.}
        \label{fig:magnetic_field_maps}
\end{figure*}

\begin{figure}
     \centering
               \begin{subfigure}[(c)]{0.2\textwidth}
         \centering
         \includegraphics[width=\textwidth]{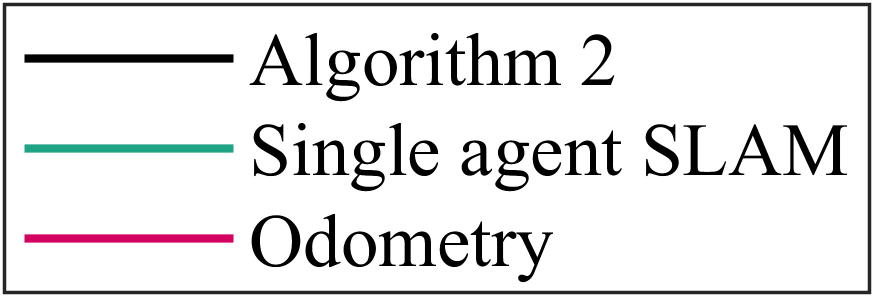} 
         \label{fig:error1}
     \end{subfigure}
     \begin{subfigure}[b]{0.45\textwidth}
         \centering
         \includegraphics[width=\textwidth]{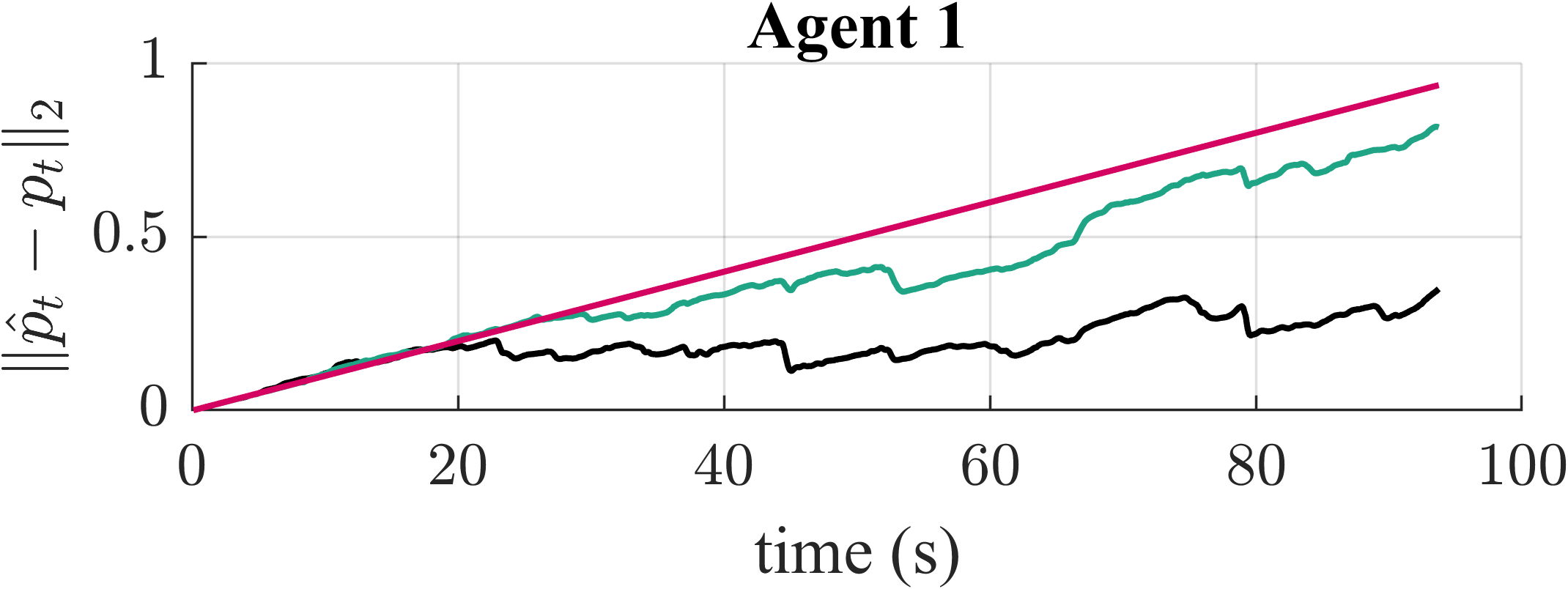}
         \label{fig:error2}
     \end{subfigure}
     \hfill
     \begin{subfigure}[b]{0.45\textwidth}
         \centering
         \includegraphics[width=\textwidth]{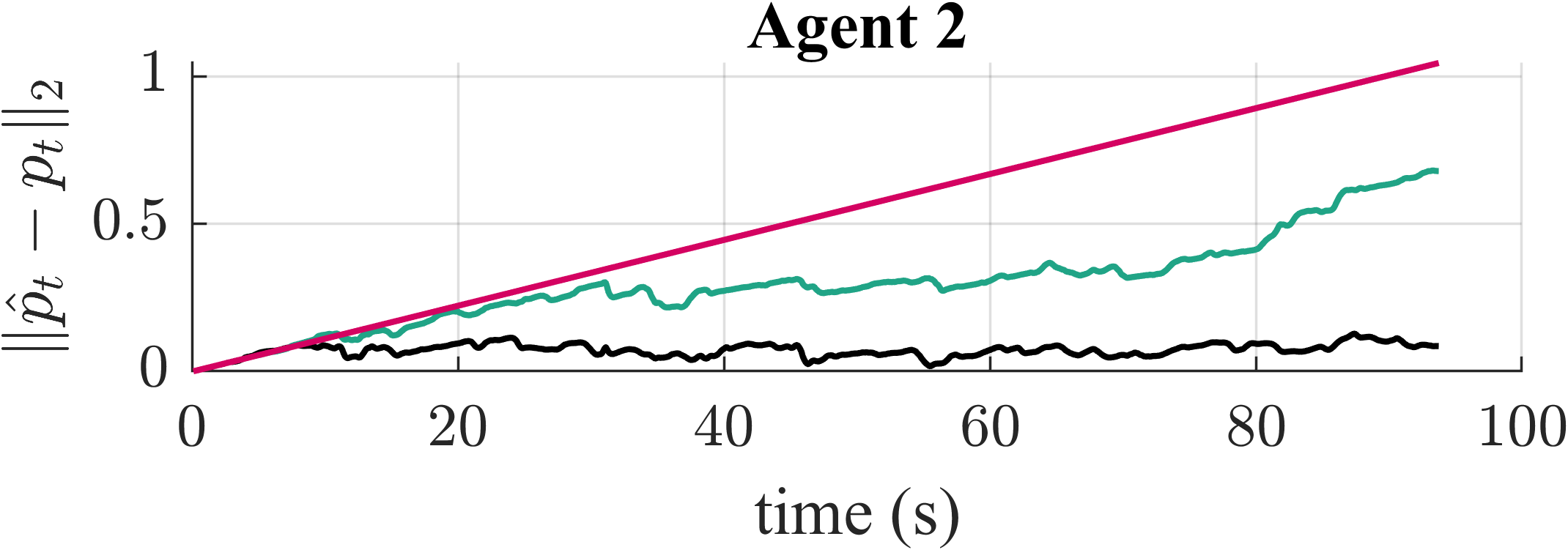}
         \label{fig:error3}
     \end{subfigure}
     \hfill
     \begin{subfigure}[(c)]{0.45\textwidth}
         \centering
         \includegraphics[width=\textwidth]{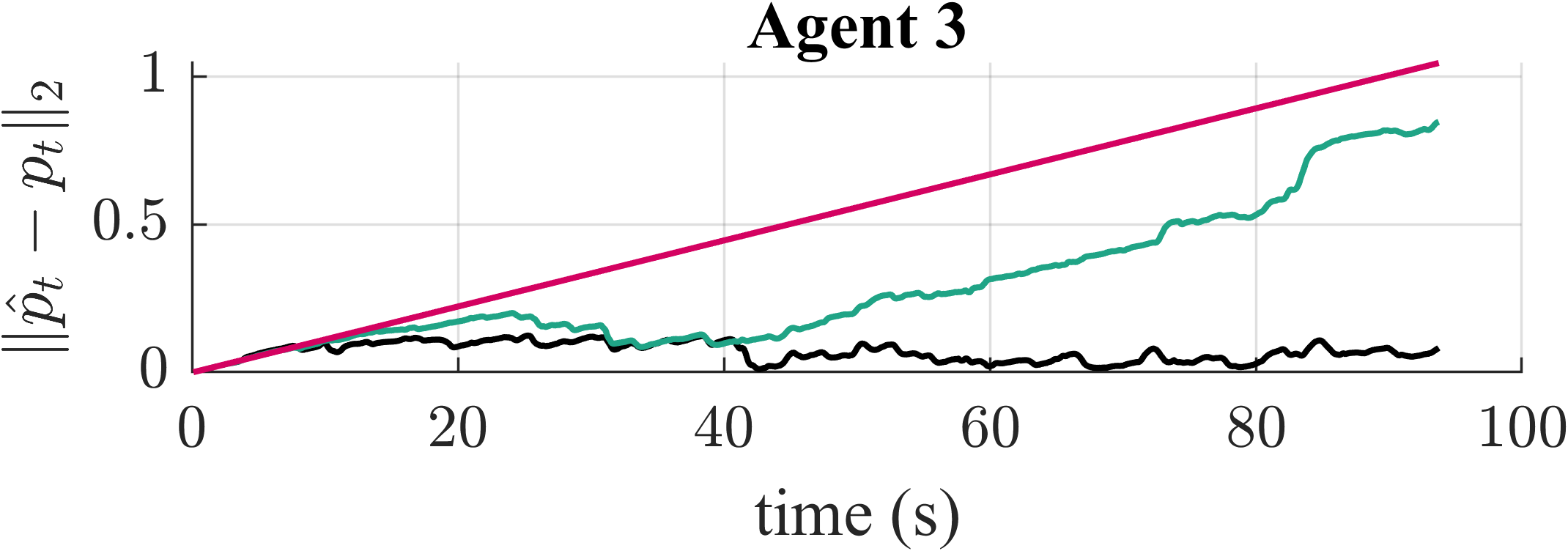} 
         \label{fig:errorslegend}
     \end{subfigure}
        \caption{Position estimation errors for three agents, relative to known ground truth position measured with optical motion capture system. The error is given as the euclidian distance between the true position $p_t$ and the estimated position $\hat{p}_t$ from Algorithm 2 with $N_c=10$ and $\alpha=0.2$, from Single agent SLAM and from integrating the pure odometry, respectively.}
        \label{fig:errors}
\end{figure}

We test the ability of our algorithm to simultaneously estimate the locations of three handheld devices containing magnetometers, by testing on data collected by three test subjects in a motion capture lab. The experimental setup is illustrated in Figure~\ref{fig:exp_setup}. Each test subject held an Xsens MTi-100 IMU, which was used to collect magnetic field measurements. The ground truth position and orientation of the IMU were recorded with an optical motion capture system. The test subjects moved sequentially in the test area to ensure marker visibility for the optical motion capture system, but we test our algorithm on the three measured trajectories as if collected simultaneously. 

The position measurements $\Delta p_{i,t}$ were simulated by first computing the difference of the recorded ground truth positions from each timestep to the next, and then adding noises of $e_{1,\text{p},t}=[0.000,\: 0.001\; 0]$, $e_{2,\text{p},t}=[-0.001,\: -0.0005,\: 0]$ and $e_{3,\text{p},t}=[0.001,\:  -0.0005,\: 0]$. These noise values were selected such that the position estimates of the agents would drift in different directions over a short timescale, making the dead-reckoning position estimates to other agents particularly poor.

The differential orientation measurements $\Delta q_{i,t}$ were simulated by computing the difference in orientation from one timestep to the next, and then adding a simulated noise sampled from a normal distribution with standard deviation $\sigma_q=1.0e-5$.
We then applied Algorithm~1 to the magnetic field measurements and the simulated odometry. The Gaussian process hyperparameters were set to $\sigma_{\text{SE}}=0.074$, $\sigma_y=0.0042$, $l_{\text{SE}}=0.86\text{m}$. The hyperparameters were selected based on an optimization of the Gaussian process likelihood, using the recorded position for all the agents as the input locations and the magnetic field norm as the output. The parameter $\sigma_p$ used in the estimation was set to $0.022$, which is two times as high as the maximum norm of the simulated noise, to make sure that the Kalman filter did not put too much trust in the odometry. To approximate the Gaussian process, $100$ basis functions were used in a domain $\Omega$ defined as the smallest cube that was no closer than $3$ meters to the closest recorded position. This is a sufficient amount of basis functions, as the approximation error between the reduced rank and the full GP in ten test points selected in random locations sampled from a uniform distribution inside the domain given all the collected measurements is lower than one measurement noise standard deviation $\sigma_y$.

The estimated trajectories using Algorithm~1 are displayed together with the learned magnetic field in Fig.~\ref{fig:magnetic_field_maps}.  The results in Figure~\ref{fig:errors} show that the EKF for a single agent improves on the position estimate for all three agents. The end-point estimation error for Single agent SLAM is $87\%$, $65\%$, and $81\%$ of the odometry error, respectively. Over time, the position estimates for Single agent SLAM are typically bounded~\cite{viset_extended_2022}, but for this example on this timescale, each agent does not have time to collect sufficient information about the magnetic field to compensate fully for the odometry drift. Even in this challenging case for magnetic field SLAM, multi-agent SLAM is able to compensate for the odometry drift. The end-point estimation errors of the position estimates from Algorithm~1 in Figure~\ref{fig:errors} are $37\%$, $8.2\%$ and $7.9\%$ compared to odometry error, for the three agents respectively.

\subsection{Testing Algorithm~2 on real magnetic field measurements with simulated odometry noise}

\begin{figure}
    \centering
    \includegraphics[width=0.45\textwidth]{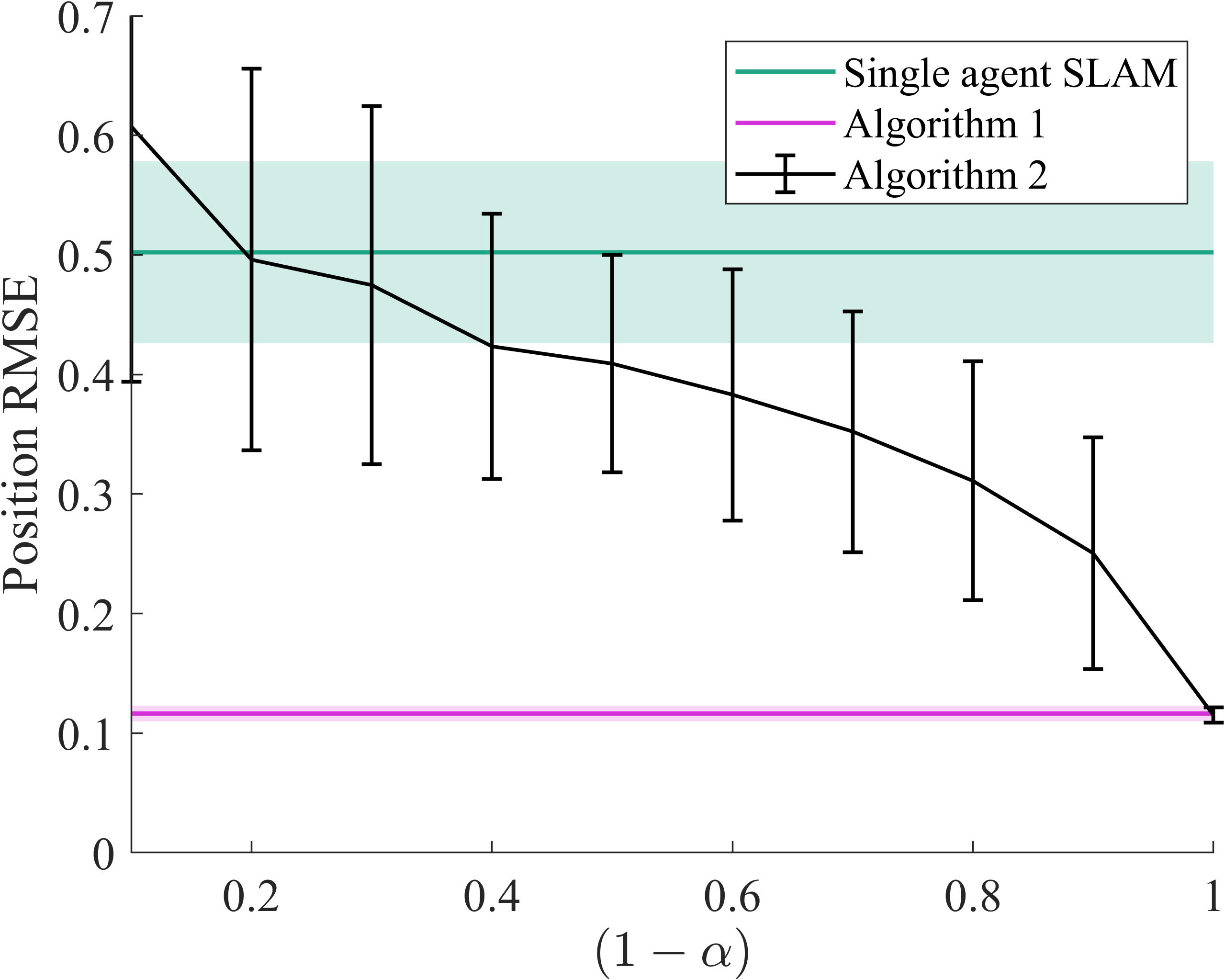}
    \caption{RMSE of the full trajectory estimate using measurements from the motion capture lab, for a range of communication failure rates $\alpha$. The error bars indicate one standard deviation after 100 Monte-Carlo repetitions. The green line marks the average deviation in the position estimate between the single-agent SLAM solution and the centralized solution after 100 Monte-Carlo repetitions, and the light green area marks the range of one standard deviation. Algorithm 2 was run with $N_c=1$.}
    \label{fig:avg_consensus_MC_reps}
\end{figure}

\begin{figure}
    \centering
    \includegraphics[width=0.45\textwidth]{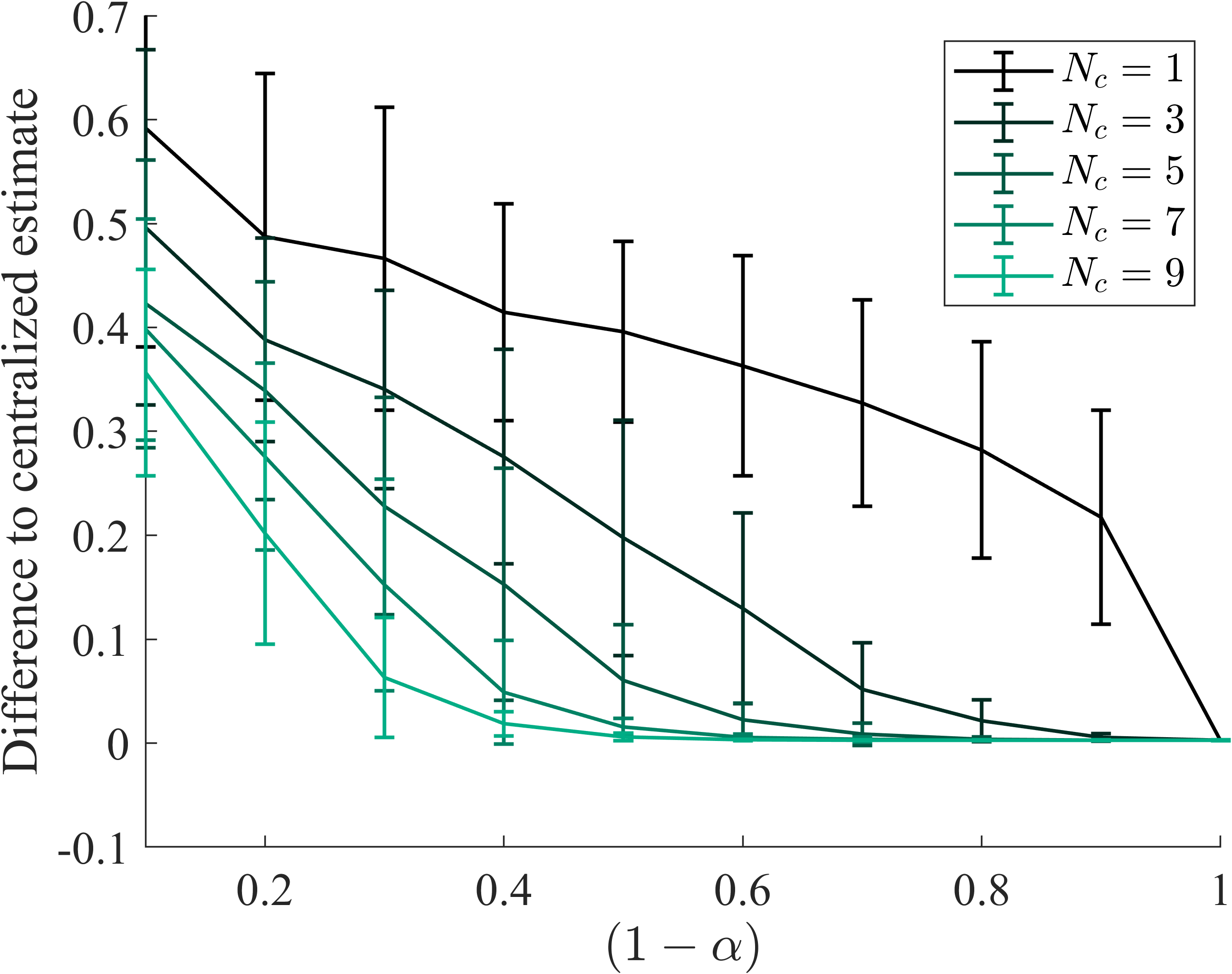}
    \caption{Deviation between the estimate from Algorithm 1 and Algorithm 2 using measurements from the motion capture lab, for a range of communication rates $\alpha$, and a range of communication steps at each iteration $N_c$. The lines connect the average results after 100 MC repetitions, and the error bars indicate one standard deviation.}
    \label{fig:avg_consensus_MC_reps_vary_Nc}
\end{figure}

We investigate the effects of varying communication failure rates $\alpha$ on the difference between the distributed estimate from Algorithm~2 and the centralized estimate from Algorithm~1. To study the most challenging case, we assume the agents have the possibility to communicate only once for each average consensus problem. By using the approximation obtained through one step of average consensus, we see in Figure~\ref{fig:avg_consensus_MC_reps} that the distributed algorithm is able to give an improved position estimate compared to single-agent magnetic field SLAM for failure rates up until $80\%$.

Each average consensus problem will give a solution that is exactly corresponding to the centralized solution when the communication failure rate is zero~\cite{olfati-saber_distributed_2005}. Otherwise, average consensus gives an approximation that converges to the true estimate as $N_c\rightarrow\infty$. The results in Fig.~\ref{fig:avg_consensus_MC_reps} confirm that the estimation error of Algorithm~\ref{alg:central} is equivalent to the estimation error of Algorithm~\ref{alg:distributed} when the dropout rate is zero. Furthermore, the results in Fig.~\ref{fig:avg_consensus_MC_reps} show that increasing the dropout rate $\alpha$, increases the estimation error of Algorithm~2. For all dropout rates of $80\%$ or lower, the resulting position estimate from Algorithm~2 is closer to the centralized solution compared to the Single-agent SLAM. The results in Fig.~\ref{fig:avg_consensus_MC_reps_vary_Nc} show that for higher $N_c$, the distributed estimate converges more rapidly to the centralized estimate as $\alpha$ increases. When the communication failure rate is zero, so for $1-\alpha=1$, we can observe that the position estimate from the distributed EKF is equivalent to the position estimate from the centralized algorithm. 

\subsection{Indoor experiment with three smartphone measurements}

\begin{figure}
    \centering
    \includegraphics[trim={135pt 60pt 115pt 200pt},clip,width=\columnwidth]{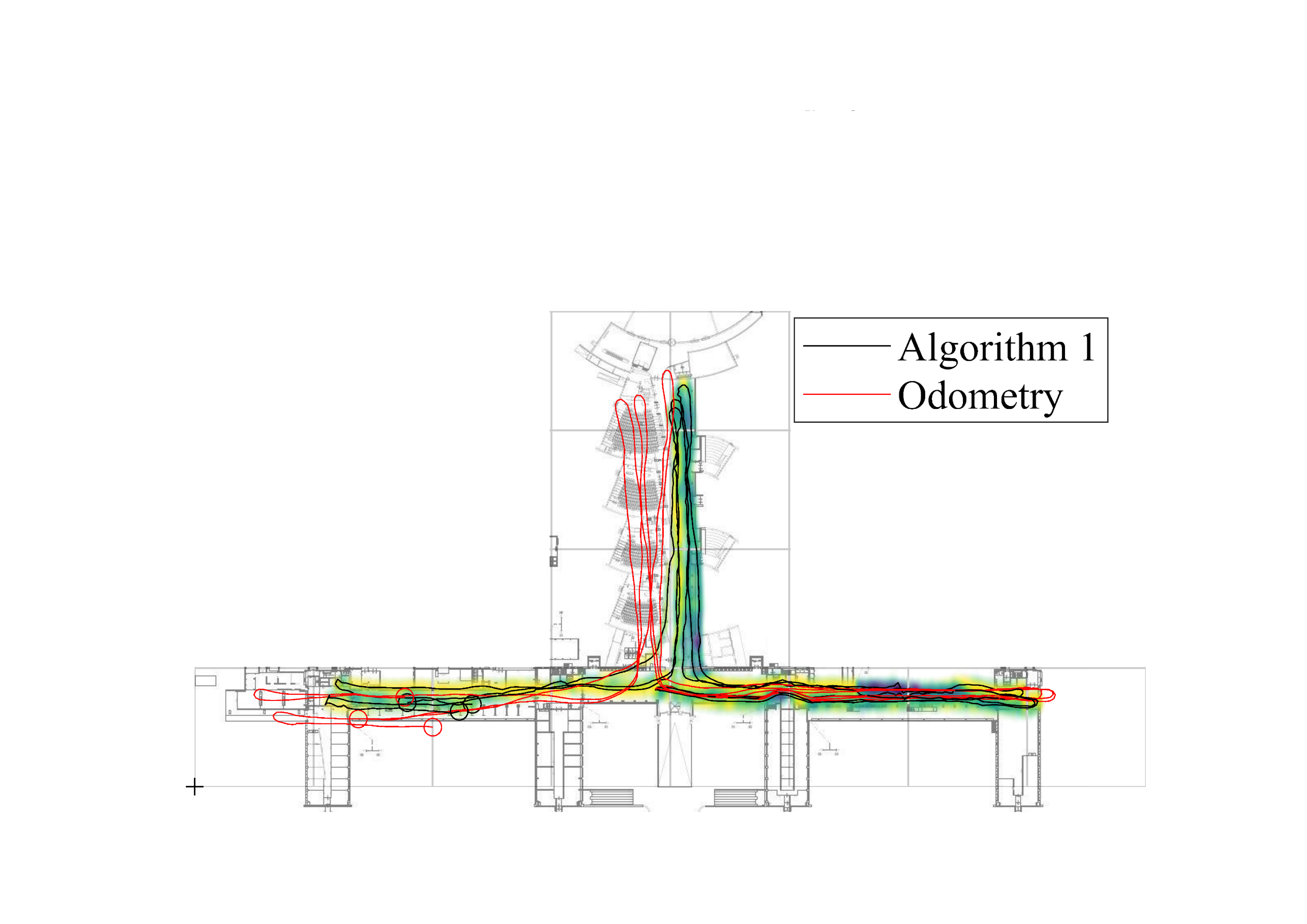}
    \caption{Learned magnetic field map and estimated trajectories for three agents in a large building.  The black circles indicate the estimated end positions of the agents using Algorithm 1, while the red circles indicate the estimated end positions of the agents using visual-inertial odometry. The color of the map is proportional to the learned intensity of the magnetic field norm, while the opacity is inversely proportional with the marginal variance.}
    \label{fig:multiagent3memagneticfield}
\end{figure}

To test our algorithm on a larger scale experiment with real odometry and magnetic field norm measurements, we collected three sequences of visual-inertial odometry and magnetic field norm measurements inside a building using Google Pixel smartphone. Google provides a platform primarily targeted at building augmented reality Apps called ARCore. Among other features, ARCore uses the phone's camera, accelerometer and gyroscope to compute a position and orientation estimate. Using a customized app, we simultaneously recorded this position and orientation estimate and the magnetometer measurements from the phone's built-in magnetometer at 200Hz. We subsequently computed the magnetic field norm using the three-component magnetic field measurements, and down-sampled all measurements to 10 Hz. Algorithm~1 was applied to these three sequences as if they were collected by three separate agents simultaneously. The algorithm was applied with the following hyperparameters: $\sigma_{\text{SE}}=7.2$, $l_{\text{SE}}=1.2\text{m}$, $\sigma_\text{y}=1.2$, $\sigma_\text{p}=0.15$, $\sigma_\text{q}=0.0001$ and with $500$ basis functions in cubic tiles of size $38 \text{m}\times38\text{m}\times 38\text{m}$. The tiles were placed with $8$ meters of overlap at the borders. The resulting visual-inertial odometry estimate of the three trajectories is displayed in Fig.~\ref{fig:multiagent3memagneticfield}. The trajectories are illustrated with respect to the floor plan of the building where they were collected. The visual-inertial odometry is initially close to the real position, but over time, it drifts away from the hallways where the measurements were collected. In the same figure, the resulting position estimate of Algorithm~1 is displayed. These estimates are closer to the hallways where the measurements were collected, and therefore likely to have higher accuracy. The magnetic field map learned collaboratively by the three agents is displayed in Fig.~\ref{fig:multiagent3memagneticfield}.

\section{Conclusion}

For multiple agents navigating in a new environment, we proposed two algorithms that allow them to collaborate about solving the simultaneous mapping and localization task. The first algorithm can be employed when a central unit has access to all measurements from all agents. The second algorithm allows for multiple agents to collaboratively approximate the estimate of the first algorithm when there is no central station that can communicate with all agents at all times. Our proposed algorithms are capable of compensating for drift also in cases where single-agent SLAM fails to do so. We presented experimental results that confirm that the centralized multi-agent SLAM algorithm obtains a higher position accuracy compared to single-agent magnetic field SLAM. For our experimental results, the second algorithm was shown to give more accurate position estimates compared to single-agent SLAM for communication drop-out rates up until $80\%$.

\bibliography{bibliography} 
\bibliographystyle{ieeetr}

\appendix

\subsection{Basis function definitions}\label{app:basis_functions}

The basis functions are defined over a finite-support cubical domain $\Omega\subset\mathbb{R}^d$, defined as $\Omega=[L_{l,1},L_{u,1}]\times [L_{l,2},L_{u,2}] \times [L_{l,3},L_{u,3}]$. The basis functions are given as
\begin{align}
    \phi_i(p)=\prod_{d=1}^3 \frac{\sqrt{2}}{\sqrt{L_{\text{u},d}-L_{\text{l},d}}} \sin\left(\frac{\pi n_{i,d}(p_d+L_{\text{l},d})}{L_{\text{u},d}-L_{\text{l},d}}\right ),\label{eq:basis_functions_closed_form}
\end{align}
where the set $(n_{i,1},n_{i,2},n_{i,3})$ is the set of three natural numbers that is different from the sets $(n_{j,1},n_{j,2},n_{j,3})$ defined for all $j<i$, that gives the corresponding value of a parameter $\lambda_i$ defined as
\begin{align}
    \lambda_{i}=\sum_{d=1}^D\left(\frac{\pi n_{i,d}}{L_{\text{u},d}-L_{\text{l},d}}\right)^2,\label{eq:eigen_values_closed_form}
\end{align}
as large as possible.
These basis functions are then used to approximate the Gaussian process prior with a parametric prior
\begin{equation}\label{eq:approximate_GP_prior}
    f\approx \Phi^\top w, \qquad w\sim\mathcal{N}(0,\Lambda),
\end{equation}
where $\Phi$ is a vector of $M$ basis functions $\phi_i:\mathbb{R}^{d}\rightarrow\mathbb{R}$, $w\in\mathbb{R}^M$ is a vector of weights, and $\Lambda$ is defined as
\begin{equation}\label{eq:red_rank_prior}
\begin{aligned}
    &\Lambda=\text{diag}\left [ S_{\text{SE}}(\sqrt{\lambda_1}),\quad\cdots,\quad S_{\text{SE}}(\sqrt{\lambda_{N_m}}) \right ],
\end{aligned}
\end{equation}
with $S_{\text{SE}}(\cdot)$ being the spectral density of the squared exponential kernel, as defined in \cite{wahlstrom_modeling_2013}. This means that the approximation of the magnetic field norm in~\eqref{eq:approximate_GP_prior} has a prior distribution that tends to~\eqref{eq:GP_prior} as $M$ goes to infinity, and the size of the domain goes to infinity~\cite{solin_explicit_2014}.

\newpage

\end{document}